\documentclass{article}
\usepackage{iclr2026_conference,times}

\usepackage{hyperref}
\usepackage{url}
\usepackage{amsmath,amssymb,amsfonts}
\usepackage{algorithmic}
\usepackage{algorithm}
\usepackage{graphicx}
\usepackage{booktabs}
\usepackage{multirow}
\usepackage{xcolor}
\usepackage{pifont}
\usepackage{tikz}
\usetikzlibrary{arrows.meta, positioning, shapes, calc}

\newcommand{\doop}{\text{do}}
\newcommand{\Pa}{\text{Pa}}
\newcommand{\E}{\mathbb{E}}

\newcommand{\Gcal}{\mathcal{G}}
\newcommand{\Scal}{\mathcal{S}}

\title{Interventional Time Series Priors for \\ Causal Foundation Models}

\author{Dennis Thumm \\
National University of Singapore \\
\texttt{dennis.thumm@u.nus.edu} \\
\And
Ying Chen \\
Department of Mathematics \\
Centre for Quantitative Finance \\
Risk Management Institute \\
National University of Singapore \\
\texttt{matcheny@nus.edu.sg}
}

\iclrfinalcopy

\track{Research}

\usepackage{booktabs}
\begin{document}

\maketitle

\begin{abstract}
Prior-data fitted networks (PFNs) have emerged as powerful foundation models for tabular causal inference, yet their extension to time series remains limited by the absence of synthetic data generators that provide interventional targets. Existing time series benchmarks generate observational data with ground-truth causal graphs but lack the interventional data required for training causal foundation models. To address this, we propose \textbf{CausalTimePrior}, a principled framework for generating synthetic temporal structural causal models (TSCMs) with paired observational and interventional time series. Our prior supports configurable causal graph structures, nonlinear autoregressive mechanisms, regime-switching dynamics, and multiple intervention types (hard, soft, time-varying). We demonstrate that PFNs trained on CausalTimePrior can perform in-context causal effect estimation on held-out TSCMs, establishing a pathway toward foundation models for time series causal inference.
\end{abstract}

\section{Introduction}

Foundation models have transformed machine learning by enabling test-time inference without task-specific training. In tabular domains, prior-data fitted networks (PFNs) achieve this by pre-training transformers on synthetic datasets sampled from a prior distribution over data-generating processes \citep{muller2022transformers, hollmann2023tabpfn}. Recent work extends PFNs to causal inference: Do-PFN \citep{robertson2025dopfn} and CausalFM \citep{ma2025causalfm} train on synthetic structural causal models (SCMs) \citep{pearl2009causality} with interventional data, enabling in-context estimation of treatment effects from observational data alone.

However, extending causal PFNs to time series faces a fundamental obstacle: the lack of suitable synthetic data generators. While benchmarks like CausalTime \citep{cheng2023causaltime}, TimeGraph \citep{ferdous2025timegraph}, and CauseMe \citep{runge2019causeme} provide time series with ground-truth causal graphs, they generate only \textit{observational} data. 
Without interventional targets, one cannot train models to predict outcomes under interventions---the core task of causal inference.

We address this gap with \textbf{CausalTimePrior}\footnote{\url{https://github.com/thummd/CausalTimePrior}}, a framework for sampling temporal SCMs (TSCMs) together with paired observational and interventional time series (see Table~\ref{tab:comparison}). 
Our contributions are (1) a \textbf{practical prior over discrete-time dynamic SCMs} \citep{boeken2024dscm} that generates paired observational and interventional time series for training causal foundation models, (2) support for \textbf{regime-switching TSCMs} with Markov-driven structural breaks and interventional data---the first generator combining regime-switching dynamics \citep{balsells2024identifiability} with interventional time series generation, and (3) \textbf{preliminary experiments} demonstrating that PFNs trained on CausalTimePrior can predict interventional outcomes given only observational context.
\section{Related Work}
\label{sec:related}

\paragraph{Time series causal discovery benchmarks.} CausalTime \citep{cheng2023causaltime} fits neural networks to real observations and extracts causal graphs via importance analysis, producing realistic data with known ground truth. TimeGraph \citep{ferdous2025timegraph} generates synthetic time series from linear and nonlinear SCMs with configurable graph properties. 
CauseMe \citep{runge2019causeme} benchmarks (including Lorenz-96 systems) and CausalRivers \citep{stein2025causalrivers}, the largest real-world benchmark (1,160 stations), offer physical ground-truth graphs.
In contrast to CausalTimePrior, these existing methods are limited to observational data without interventions.

\paragraph{Generators with interventional support.} We identified only three frameworks supporting temporal interventions, each with limitations.
\textbf{CAnDOIT} \citep{castri2024candoit} generates time-lagged SCMs with hard interventions on \textit{known, single-node} targets. It supports nonlinear mechanisms but only static intervention values.
\textbf{TECDI/RealTCD} \citep{li2023tecdi, li2024realtcd} use structural VAR models with soft (TECDI) or hard (RealTCD) interventions. RealTCD handles unknown targets but is limited to linear mechanisms.
\textbf{CaTSG} \citep{xia2025catsg} implements $\doop$-calculus \citep{pearl2016causal} via diffusion models with a causal score function. While CaTSG implements interventions via learned diffusion models requiring training on specific datasets, CausalTimePrior generates interventional data analytically from explicit structural equations, enabling fast prior sampling without a separate generative model.

\begin{table}[t]
\caption{Comparison of time series causal data generators. CausalTimePrior is the first to support regime-switching dynamics (changing causal structures over time).}
\label{tab:comparison}
\centering
\small
\vskip 0.1in
\begin{tabular}{@{}lcccc@{}}
\toprule
\textbf{Generator} & \textbf{Interventions} & \textbf{Nonlinear SCMs} & \textbf{Time-varying Interventions} & \textbf{Regime-switch} \\
\midrule
CausalTime & \textcolor{red}{\ding{55}} & \textcolor{green!60!black}{\ding{51}} & --- & \textcolor{red}{\ding{55}} \\
TimeGraph & \textcolor{red}{\ding{55}} & \textcolor{green!60!black}{\ding{51}} & --- & \textcolor{red}{\ding{55}} \\
CAnDOIT & \textcolor{green!60!black}{\ding{51}} Hard & \textcolor{green!60!black}{\ding{51}} & \textcolor{red}{\ding{55}} & \textcolor{red}{\ding{55}} \\
TECDI & \textcolor{green!60!black}{\ding{51}} Soft & \textcolor{red}{\ding{55}} & \textcolor{red}{\ding{55}} & \textcolor{red}{\ding{55}} \\
CaTSG & \textcolor{green!60!black}{\ding{51}} $\doop$-calculus & \textcolor{green!60!black}{\ding{51}} & \textcolor{red}{\ding{55}} & \textcolor{red}{\ding{55}} \\
\midrule
\textbf{CausalTimePrior} & \textcolor{green!60!black}{\ding{51}} All & \textcolor{green!60!black}{\ding{51}} & \textcolor{green!60!black}{\ding{51}} & \textcolor{green!60!black}{\ding{51}} \\
\bottomrule
\end{tabular}
\end{table}

\paragraph{Causal PFNs for tabular data.} Do-PFN \citep{robertson2025dopfn} pre-trains transformers on synthetic SCMs to predict conditional interventional distributions (CIDs) without knowing the causal graph. CausalFM \citep{ma2025causalfm} formalizes Bayesian priors over SCMs for back-door, front-door, and instrumental variable settings. Both demonstrate strong performance on i.i.d.\ tabular data but do not address temporal dependencies. Related work on counterfactual time series estimation includes CRN \citep{bica2020crn} and the Causal Transformer \citep{melnychuk2022causal}, which estimate individualized treatment effects over time but require per-dataset training rather than in-context learning.

\paragraph{Time series foundation models.} Recent work has explored zero-shot forecasting via synthetic pre-training. ForecastPFN \citep{dooley2023forecastpfn} and TimePFN \citep{taga2025timepfn} pre-train transformers on synthetic data for multivariate forecasting. TempoPFN \citep{moroshan2025tempopfn} pre-trains linear Recurrent Neural Networks (RNNs) for univariate forecasting, using diverse synthetic generators including Stochastic Differential Equations (SDEs), Gaussian processes, and causal kernels (CauKer) \citep{xie2025cauker}. While CauKer generator produces multivariate SCM-based time series, it lacks temporal lag structures and intervention support. These models target prediction rather than causal inference.

\paragraph{Regime-switching dynamics.} Switching Dynamical Systems (SDSs) with Markov Switching Models provide identifiability theory for regime-dependent causal discovery \citep{balsells2024identifiability}, but focus on inference from observational data without generating interventional datasets for training foundation models.

\section{CausalTimePrior}
\label{sec:method}

We define a prior $\Pi$ over temporal structural causal models that generates paired observational and interventional time series suitable for training causal foundation models (see Algorithm \ref{alg:prior}).

\subsection{Temporal Structural Causal Models}

Following the Dynamic Structural Causal Model (DSCM) framework \citep{boeken2024dscm}, we consider the discrete-time acyclic case. A temporal SCM (TSCM) $\Scal = (\Gcal, \mathbf{F}, P_\epsilon)$ consists of:
\begin{itemize}
    \item A \textbf{time-lagged DAG} $\Gcal = (G_0, G_1, \ldots, G_K)$ where $G_0 \in \{0,1\}^{N \times N}$ encodes instantaneous (intra-slice) edges and $G_k$ encodes edges from time $t-k$ to $t$ for lags $k \in \{1, \ldots, K\}$.
    \item \textbf{Structural equations} $\mathbf{F} = \{f_i\}_{i=1}^N$ where:
    \begin{equation}
        X_t^{(i)} = f_i\left(\Pa_{\Gcal}(X_t^{(i)})\right) + \epsilon_t^{(i)}, \quad \epsilon_t^{(i)} \sim P_{\epsilon}^{(i)}
    \end{equation}
    with parents $\Pa_{\Gcal}(X_t^{(i)}) = \{X_{t-k}^{(j)} : G_k[j,i] = 1, k \in \{0,\ldots,K\}\}$.
\end{itemize}

\subsection{Prior Distribution over TSCMs}


\paragraph{Graph prior $\Pi_\Gcal$.} We sample the number of variables $N \sim \text{Uniform}(3, N_{\max})$, maximum lag $K \sim \text{Uniform}(1, K_{\max})$, and edge probability $p \sim \text{Beta}(\alpha, \beta)$. Instantaneous edges $G_0$ are sampled from an Erd\H{o}s-R\'{e}nyi model \citep{erdos1960evolution} with acyclicity enforced via topological ordering. Lagged edges $G_k$ are sampled independently with probability decaying as $p \cdot \gamma^k$ for persistence factor $\gamma \in (0,1]$.

\paragraph{Mechanism prior $\Pi_\mathbf{F}$.} 
We sample mechanisms from multiple families. For simple mechanisms:
\begin{equation}
    f_i(\mathbf{x}) = \sum_{j \in \Pa(i)} w_{ij} \cdot \phi_{ij}(x_j) + b_i
\end{equation}
where weights $w_{ij} \sim \mathcal{N}(0, \sigma_w^2)$, biases $b_i \sim \mathcal{N}(0, \sigma_b^2)$, and $\phi_{ij}$ is sampled uniformly from $\{\text{id}, \sin, \cos, \tanh, |\cdot|, (\cdot)^2, \exp(-|\cdot|)\}$. The diversity of activation functions ensures the prior covers a wide range of nonlinear temporal dynamics.

\paragraph{Noise prior $\Pi_\epsilon$.} Noise distributions are sampled per variable from $\{\mathcal{N}(0, \sigma^2), \text{Uniform}(-a, a), \text{Laplace}(0, b)\}$ with scale parameters drawn from suitable hyperpriors.

\subsection{Intervention Types}

Given a sampled TSCM $\Scal$, we generate interventional data by modifying structural equations \citep{eberhardt2007interventions}. Let $I \subseteq \{1, \ldots, N\}$ denote intervention targets and $t_I \subseteq \{1, \ldots, T\}$ the intervention times. We provide examples for each intervention type in Appendix \ref{app:intervention_examples}.

\paragraph{Hard interventions.} ($\doop$-operator) Replace $X_t^{(i)} := c$ for $i \in I, t \in t_I$, severing incoming edges.

\paragraph{Soft interventions.} Perturb the mechanism: $X_t^{(i)} = f_i(\Pa(X_t^{(i)})) + \delta_i + \epsilon_t^{(i)}$ for shift $\delta_i \sim \mathcal{N}(\mu_\delta, \sigma_\delta^2)$.

\paragraph{Time-varying interventions.} Set $X_t^{(i)} := c(t)$ where $c(t)$ follows a specified profile (step, ramp, sinusoidal, or sampled trajectory) \citep{hernan2020causal}.

\subsection{Data Generation Pipeline}

For each training example, we:
\begin{enumerate}
    \item Sample $\Scal \sim \Pi$ (graph, mechanisms, noise).
    \item Sample intervention specification: targets $I$, times $t_I$, type, and values.
    \item Generate \textbf{observational} series $\mathbf{X}^{\text{obs}}_{1:T}$ by forward simulation.
    \item Generate \textbf{interventional} series $\mathbf{X}^{\text{int}}_{1:T}$ under $\doop(X^{(I)}_{t_I} = c)$.
    \item Form training tuple: $(\mathbf{X}^{\text{obs}}_{1:T}, I, t_I, c, Y^{\text{int}}_\tau)$ where $Y^{\text{int}}_\tau$ is the outcome variable at target time $\tau$ under intervention.
\end{enumerate}

\subsection{Regime-Switching Priors}

Real-world time series often exhibit structural breaks where causal relationships change \citep{rahmani2025fantom}. We extend our prior to regime-switching TSCMs, following the Markov Switching Model framework \citep{balsells2024identifiability}:
\begin{equation}
    X_t^{(i)} = f_i^{(d_t)}\left(\Pa_{\Gcal^{(d_t)}}(X_t^{(i)})\right) + \epsilon_t^{(i)}, \quad d_t \sim \text{Markov}(\mathbf{P})
\end{equation}
where $d_t \in \{1, \ldots, R\}$ indexes the active regime with transition matrix $\mathbf{P}$. Each regime has its own causal graph $\Gcal^{(r)}$ and mechanisms $\mathbf{F}^{(r)}$. Regime transitions follow a sticky Markov chain ($P_{ii} \approx 0.9$) to model persistent causal structures. In our experiments, 15\% of training TSCMs are regime-switching with $R \in \{2, 3\}$ regimes. Combined with interventional data generation, this enables training PFNs that can reason about interventions under time-varying causal structures.

\subsection{Design Choices and Assumptions}

\paragraph{Complexity, identifiability, and causality.} Adding lags increases model complexity \emph{linearly}, not exponentially: each lag $k$ contributes $O(N)$ weight parameters per node, and edge density decays geometrically as $p_k = p \cdot \gamma^k$ ($\gamma = 0.7$), so distant lags are increasingly sparse. Identifiability holds per time step: within each $t$, variables are evaluated in topological order of $G_0$, so instantaneous parents are already computed, while lagged parents $X^{(j)}_{t-k}$ are fixed from prior steps. This naturally resolves instantaneous and temporal causality through a single forward simulation---$G_0$ captures same-timestep effects (akin to imputation), while $G_1, \ldots, G_K$ capture lagged effects (akin to forecasting), with temporal ordering preventing cycles by construction.

\paragraph{Noise assumptions and prediction horizon.} Noise is \emph{additive} and \emph{exogenous}: $\epsilon_t^{(i)}$ is added after the nonlinear activation and sampled independently of parent values. It is also \emph{Markovian}---i.i.d.\ across time---implying that the causal graph fully mediates all temporal dependencies; extending to non-Markovian (temporally correlated) noise would model latent confounders persisting across time and is an important future direction. At training time, the PFN queries outcomes 0--5 steps after intervention onset (70\% downstream queries at 1--5 steps, 30\% instantaneous), concentrating learning on the short-range causal propagation window; see Appendix~\ref{app:assumptions} for further discussion.

\section{Experiments}
\label{sec:experiments}

\paragraph{Prior Validation.}
We validate CausalTimePrior by analyzing a dataset of 100K generated TSCMs.
(1) \textbf{Structural diversity}: the prior generates TSCMs with $N \in [3, 10]$ variables, $K \in [1, 3]$ lags, and $T = 50$ time steps, including 70\% diverse nonlinear TSCMs, 15\% chain structures, and 15\% regime-switching TSCMs. Erd\H{o}s-R\'{e}nyi sampling with varied edge probabilities implicitly produces canonical causal motifs (confounders, mediators, colliders) as subgraphs.
(2) \textbf{Stability}: 0\% divergence rate (no NaN/Inf values) across all 100K samples, achieved through value clipping and careful mechanism parameterization.
(3) \textbf{Intervention coverage}: hard, soft, and time-varying interventions with mean effect size of 17.98 (std 53.93), demonstrating substantial variability in intervention magnitudes across types.
(4) \textbf{Paired data quality}: observational and interventional series maintain similar statistics (obs: $\mu=46.78$, $\sigma=242.85$; int: $\mu=41.56$, $\sigma=228.52$), confirming that interventions produce realistic counterfactual outcomes rather than out-of-distribution artifacts. 
Example visualization and full distributions of prior properties are shown in Appendix~\ref{app:visualizations} and \ref{app:prior_distributions}.

\paragraph{Proof-of-Concept PFN.}
As a preliminary demonstration, we train a simple 2-layer GRU-based PFN (128 hidden dim, 11 min on CPU) on 100K TSCMs from CausalTimePrior and evaluate on 1,000 held-out TSCMs. The model learns to distinguish causal from non-causal queries: Pred/GT ratio of 0.95 for intervened queries vs.\ 0.46 for non-causal queries (Table~\ref{tab:results}), and achieves comparable RMSE to per-dataset VAR baselines without per-sample fitting (Table~\ref{tab:baselines}). Implementation details, full results, baselines, a shuffled-intervention control experiment, ablations, generalizations, and an example are in Appendix~\ref{app:details}, \ref{app:results}, \ref{app:ablation}, \ref{app:ood}, and \ref{app:causal_example}.

\section{Conclusion}

CausalTimePrior addresses a critical gap in time series causal inference: the absence of synthetic generators with interventional data for training foundation models.  
By combining diverse temporal generators with principled intervention logic, it yields a prior over TSCMs with diverse intervention types. 
Our preliminary results suggest that PFNs trained on this prior can perform in-context causal effect estimation, opening a pathway toward foundation models for time series causality.

\paragraph{Limitations and future work.}
The framework currently assumes Markovian noise and discrete-time dynamics; extensions to non-Markovian confounding and continuous-time processes are important future directions. Our Erd\H{o}s-R\'{e}nyi graph prior implicitly covers canonical causal structures (confounders, mediators, colliders) but does not explicitly stratify over them as Do-PFN does for tabular settings; adding structured temporal motifs could improve coverage. The prior has not been validated against real-world causal time series distributions. We plan to
(1) scale training to larger models with explicit canonical structure sampling, (2) incorporate continuous-time dynamics, and (3) benchmark on semi-synthetic datasets derived from real observational data.

\subsubsection*{Acknowledgments}
The authors are grateful for valuable discussions with Jake Roberston, Frank Hutter, and the Prior Labs team at the EurIPS'25 Workshop on AI for Tabular Data.

\bibliography{iclr2026/tsalm}

@inproceedings{bengio20009curriculum,
author = {Bengio, Yoshua and Louradour, J\'{e}r\^{o}me and Collobert, Ronan and Weston, Jason},
title = {Curriculum learning},
year = {2009},
isbn = {9781605585161},
publisher = {Association for Computing Machinery},
address = {New York, NY, USA},
url = {https://doi.org/10.1145/1553374.1553380},
doi = {10.1145/1553374.1553380},
booktitle = {Proceedings of the 26th Annual International Conference on Machine Learning},
pages = {41–48},
numpages = {8},
location = {Montreal, Quebec, Canada},
series = {ICML '09}
}

@article{peters2014causal,
  title={Causal discovery with continuous additive noise models},
  author={Peters, Jonas and Mooij, Joris M and Janzing, Dominik and Sch{\"o}lkopf, Bernhard},
  journal={Journal of Machine Learning Research},
  year={2014}
}

@inproceedings{kidger2021neural,
  title={Neural {SDE}s as Infinite-Dimensional {GAN}s},
  author={Kidger, Patrick and Foster, James and Li, Xuechen and Lyons, Terry},
  booktitle={International Conference on Machine Learning},
  pages={5453--5463},
  year={2021},
  organization={PMLR}
}

@inproceedings{chen2018neural,
  title={Neural Ordinary Differential Equations},
  author={Chen, Ricky T. Q. and Rubanova, Yulia and Bettencourt, Jesse and Duvenaud, David},
  booktitle={Advances in Neural Information Processing Systems},
  volume={31},
  year={2018}
}

@book{kloeden1992numerical,
  title={Numerical Solution of Stochastic Differential Equations},
  author={Kloeden, Peter E. and Platen, Eckhard},
  year={1992},
  publisher={Springer-Verlag},
  address={Berlin}
}

@inproceedings{
manten2025signature,
title={Signature Kernel Conditional Independence Tests in Causal Discovery for Stochastic Processes},
author={Georg Manten and Cecilia Casolo and Emilio Ferrucci and S{\o}ren Wengel Mogensen and Cristopher Salvi and Niki Kilbertus},
booktitle={The Thirteenth International Conference on Learning Representations},
year={2025},
url={https://openreview.net/forum?id=Nx4PMtJ1ER}
}

@conference{thumm2025towards,
    author = {Thumm, Dennis and Mijares, Luis Ontaneda},
    booktitle = {ICAIF 2025 Workshop on Rethinking Financial Time-Series},
    title = {Towards Causal Market Simulators},
    year = {2025},
    address   = {Singapore},
    url = {https://icaif-25-rtfs.github.io/}
}

@book{hernan2020causal,
  title={Causal Inference: What If},
  author={Hern{\'a}n, Miguel A and Robins, James M},
  publisher={Chapman \& Hall/CRC},
  year={2020},
  isbn={9780367583421},
  url={https://miguelhernan.org/whatifbook/}
}

@article{eberhardt2007interventions,
  title={Interventions and causal inference},
  author={Eberhardt, Frederick and Scheines, Richard},
  journal={Philosophy of science},
  volume={74},
  number={5},
  pages={981--995},
  year={2007},
  publisher={Cambridge University Press}
}

@article{erdos1960evolution,
  title={On the evolution of random graphs},
  author={Erd{\H{o}}s, Paul and R{\'e}nyi, Alfr{\'e}d},
  journal={Publicationes Mathematicae},
  volume={5},
  pages={17--61},
  year={1960}
}

@inproceedings{balsells2024identifiability,
  title={On the Identifiability of Switching Dynamical Systems},
  author={Balsells-Rodas, Carles and Wang, Yixin and Li, Yingzhen},
  booktitle={International Conference on Machine Learning},
  pages={2639--2672},
  year={2024},
  organization={PMLR}
}

@article{xie2025cauker,
  publtype={informal},
  author={Shifeng Xie and Vasilii Feofanov and Marius Alonso and Ambroise Odonnat and Jianfeng Zhang and Themis Palpanas and Ievgen Redko},
  title={CauKer: classification time series foundation models can be pretrained on synthetic data only},
  year={2025},
  month={August},
  cdate={1754006400000},
  journal={CoRR},
  volume={abs/2508.02879},
  url={https://doi.org/10.48550/arXiv.2508.02879}
}

@article{dooley2023forecastpfn,
  title={Forecastpfn: Synthetically-trained zero-shot forecasting},
  author={Dooley, Samuel and Khurana, Gurnoor Singh and Mohapatra, Chirag and Naidu, Siddartha V and White, Colin},
  journal={Advances in Neural Information Processing Systems},
  volume={36},
  pages={2403--2426},
  year={2023}
}

@inproceedings{taga2025timepfn,
  title={TimePFN: Effective multivariate time series forecasting with synthetic data},
  author={Taga, Ege Onur and Ildiz, Muhammed Emrullah and Oymak, Samet},
  booktitle={Proceedings of the AAAI Conference on Artificial Intelligence},
  volume={39},
  number={19},
  pages={20761--20769},
  year={2025}
}

@inproceedings{boeken2024dscm,
  title={Dynamic Structural Causal Models},
  author={Boeken, Philip and Mooij, Joris M.},
  booktitle={UAI 2024 Workshop on Causal Inference for Time Series (CI4TS)},
  year={2024}
}

@book{pearl2016causal,
  title={Causal inference in statistics: A primer},
  author={Pearl, Judea and Glymour, Madelyn and Jewell, Nicholas P},
  year={2016},
  publisher={John Wiley \& Sons}
}

@inproceedings{muller2022transformers,
  title={Transformers can do {B}ayesian inference},
  author={M{\"u}ller, Samuel and Hollmann, Noah and Arango, Sebastian Pineda and Grabocka, Josif and Hutter, Frank},
  booktitle={International Conference on Learning Representations},
  year={2022}
}

@inproceedings{hollmann2023tabpfn,
  title={{TabPFN}: A transformer that solves small tabular classification problems in a second},
  author={Hollmann, Noah and M{\"u}ller, Samuel and Eggensperger, Katharina and Hutter, Frank},
  booktitle={International Conference on Learning Representations},
  year={2023}
}

@inproceedings{robertson2025dopfn,
  title={Do-PFN: In-Context Learning for Causal Effect Estimation},
  author={Robertson, Jake and Reuter, Arik and Guo, Siyuan and Hollmann, Noah and Hutter, Frank and Sch{\"o}lkopf, Bernhard},
  booktitle={1st ICML Workshop on Foundation Models for Structured Data},
  year = {2025}
}

@article{ma2025causalfm,
  title={Foundation Models for Causal Inference via Prior-Data Fitted Networks},
  author={Ma, Yuchen and Frauen, Dennis and Javurek, Emil and Feuerriegel, Stefan},
  journal={arXiv preprint arXiv:2506.10914},
  year={2025}
}

@inproceedings{cheng2023causaltime,
  title={{CausalTime}: Realistically generated time-series for benchmarking of causal discovery},
  author={Cheng, Yuxiao and Yang, Ziqian and Chen, Xu and Li, Jiecheng and Yan, Junchi},
  booktitle={International Conference on Learning Representations},
  year={2024}
}

@inproceedings{ferdous2025timegraph,
  title={Timegraph: Synthetic benchmark datasets for robust time-series causal discovery},
  author={Ferdous, Muhammad Hasan and Hossain, Emam and Gani, Md Osman},
  booktitle={Proceedings of the 31st ACM SIGKDD Conference on Knowledge Discovery and Data Mining V. 2},
  pages={5425--5435},
  year={2025}
}

@article{runge2019causeme,
  title={Inferring causation from time series in {E}arth system sciences},
  author={Runge, Jakob and Bathiany, Sebastian and Bollt, Erik and Camps-Valls, Gustau and Coumou, Dim and Deyle, Ethan and Glymour, Clark and Kretschmer, Marlene and Mahecha, Miguel D and Mu{\~n}oz-Mar{\'\i}, Jordi and others},
  journal={Nature Communications},
  volume={10},
  number={1},
  pages={2553},
  year={2019}
}

@inproceedings{stein2025causalrivers,
  title={CausalRivers-Scaling up benchmarking of causal discovery for real-world time-series},
  author={Stein, Gideon and Shadaydeh, Maha and Blunk, Jan and Penzel, Niklas and Denzler, Joachim},
  booktitle={The Thirteenth International Conference on Learning Representations},
  year = {2025}
}

@article{castri2024candoit,
  title={{CAnDOIT}: Causal discovery with observational and interventional data from time series},
  author={Castri, Luca and Mghames, Sariah and Hanheide, Marc and Bellotto, Nicola},
  journal={Advanced Intelligent Systems},
  year={2024},
  publisher={Wiley}
}

@inproceedings{li2023tecdi,
  title={Causal discovery in temporal domain from interventional data},
  author={Li, Peiwen and Meng, Yuan and Wang, Xin and Shen, Fang and Li, Yue and Wang, Jialong and Zhu, Wenwu},
  booktitle={Proceedings of the 32nd ACM International Conference on Information and Knowledge Management},
  pages={1306--1315},
  year={2023}
}

@inproceedings{li2024realtcd,
  title={Realtcd: Temporal causal discovery from interventional data with large language model},
  author={Li, Peiwen and Wang, Xin and Zhang, Zeyang and Meng, Yuan and Shen, Fang and Li, Yue and Wang, Jialong and Li, Yang and Zhu, Wenwu},
  booktitle={Proceedings of the 33rd ACM International Conference on Information and Knowledge Management},
  pages={4669--4677},
  year={2024}
}

@article{xia2025catsg,
  title={Causal Time Series Generation via Diffusion Models},
  author={Xia, Yutong and Xu, Chang and Liang, Yuxuan and Wen, Qingsong and Zimmermann, Roger and Bian, Jiang},
  journal={arXiv preprint arXiv:2509.20846},
  year={2025}
}

@inproceedings{rahmani2025fantom,
  title={{FANTOM}: Temporal causal discovery with regime-switching normalizing flows},
  author={Rahmani, Hamed and others},
  booktitle={International Conference on Learning Representations},
  year={2025}
}

@inproceedings{melnychuk2022causal,
  title={Causal transformer for estimating counterfactual outcomes},
  author={Melnychuk, Valentyn and Frauen, Dennis and Feuerriegel, Stefan},
  booktitle={International Conference on Machine Learning},
  pages={15293--15329},
  year={2022},
  organization={PMLR}
}

@article{runge2020pcmci,
  title={Discovering contemporaneous and lagged causal relations in autocorrelated nonlinear time series datasets},
  author={Runge, Jakob},
  journal={Proceedings of the Conference on Uncertainty in Artificial Intelligence},
  pages={1388--1397},
  year={2020}
}

@book{pearl2009causality,
  title={Causality: Models, reasoning, and inference},
  author={Pearl, Judea},
  year={2009},
  publisher={Cambridge University Press}
}

@inproceedings{bica2020crn,
  title={Estimating counterfactual treatment outcomes over time through adversarially balanced representations},
  author={Bica, Ioana and Alaa, Ahmed M and Jordon, James and van der Schaar, Mihaela},
  booktitle={International Conference on Learning Representations},
  year={2020}
}

@inproceedings{moroshan2025tempopfn,
  title={{TempoPFN}: Synthetic Pre-Training of Linear {RNN}s for Zero-Shot Time Series Forecasting},
  author={Moroshan, Vladyslav and Siems, Julien and Zela, Arber and Carstensen, Timur and Hutter, Frank},
  booktitle={NeurIPS 2025 Workshop on AI for Tabular Data},
  year={2025}
}
\bibliographystyle{iclr2026_conference}

\newpage

\appendix

\section{CausalTimePrior Algorithm}
\label{app:algorithm}

Algorithm~\ref{alg:prior} formalizes the CausalTimePrior sampling procedure for generating paired observational and interventional time series from TSCMs.

\begin{algorithm}[h]
\caption{CausalTimePrior Sampling}
\label{alg:prior}
\begin{algorithmic}[1]
\STATE \textbf{Input:} Prior hyperparameters $\Pi = (\Pi_\Gcal, \Pi_\mathbf{F}, \Pi_\epsilon)$, sequence length $T$
\STATE \textbf{Output:} Observational series $\mathbf{X}^{\text{obs}}_{1:T}$, interventional series $\mathbf{X}^{\text{int}}_{1:T}$, intervention spec $I, t_I, c$
\STATE
\STATE // Sample TSCM
\STATE $N \sim \text{Uniform}(3, N_{\max})$, $K \sim \text{Uniform}(1, K_{\max})$
\STATE $p \sim \text{Beta}(\alpha, \beta)$
\STATE Sample $G_0 \in \{0,1\}^{N \times N}$ (acyclic via topological ordering)
\FOR{$k = 1$ to $K$}
    \STATE Sample $G_k$ with edge probability $p \cdot \gamma^k$
\ENDFOR
\STATE $\Gcal \gets (G_0, G_1, \ldots, G_K)$
\STATE
\FOR{$i = 1$ to $N$}
    \STATE Sample mechanism $f_i \sim \Pi_\mathbf{F}$ (nonlinear autoregressive)
    \STATE Sample noise $P_\epsilon^{(i)} \sim \Pi_\epsilon$
\ENDFOR
\STATE $\Scal \gets (\Gcal, \{f_i\}_{i=1}^N, P_\epsilon)$
\STATE
\STATE // Sample intervention specification
\STATE Sample targets $I \subseteq \{1, \ldots, N\}$
\STATE Sample times $t_I \subseteq \{1, \ldots, T\}$
\STATE Sample type $\in \{\text{hard}, \text{soft}, \text{time-varying}\}$
\STATE Sample value(s) $c$ or $c(t)$
\STATE
\STATE // Generate paired time series
\STATE $\mathbf{X}^{\text{obs}}_{1:T} \gets$ Forward simulation of $\Scal$
\STATE $\mathbf{X}^{\text{int}}_{1:T} \gets$ Forward simulation of $\Scal$ under $\doop(X^{(I)}_{t_I} = c)$
\STATE
\RETURN $\mathbf{X}^{\text{obs}}_{1:T}$, $\mathbf{X}^{\text{int}}_{1:T}$, $(I, t_I, c)$
\end{algorithmic}
\end{algorithm}

\paragraph{Continuous-time extension.}
CausalTimePrior currently generates discrete-time SCMs (Algorithm~\ref{alg:prior}), but our autoregressive mechanisms have a natural continuous-time interpretation via the Euler-Maruyama discretization \citep{kloeden1992numerical}.
Recent work on causal discovery in continuous-time SDEs \citep{manten2025signature} motivates extending CausalTimePrior to generate interventional data from SDE-based causal models.
Consider a causal Ornstein-Uhlenbeck process $dX_t = \theta(\mu - X_t)\,dt + \sigma_w\,dW_t$; applying Euler-Maruyama with step $\Delta t$ yields:
\begin{equation}
    x_{t+1} = \underbrace{(1 - \theta\Delta t)}_{c_2} x_t + \underbrace{\theta\mu\Delta t}_{c_1} + \underbrace{\sigma_w\sqrt{\Delta t}}_{c_3} Z_t, \quad Z_t \sim \mathcal{N}(0,1)
\end{equation}
which is precisely the AR(1) form our mechanism prior generates. This means each sampled discrete-time SCM can be viewed as an Euler-Maruyama discretization of a continuous-time causal SDE system \citep{thumm2025towards}. A natural extension is to sample continuous-time mechanisms directly---e.g., via Neural ODEs \citep{chen2018neural} or Neural SDEs \citep{kidger2021neural}---and discretize at variable time steps, enabling the prior to generate irregularly-sampled interventional time series.

\section{Prior Assumptions and Limitations}
\label{app:assumptions}

This section details the modeling assumptions underlying CausalTimePrior and their implications for identifiability and generalization.

\paragraph{Effect propagation and decay.}
Causal effects in the prior decay through three mechanisms: (1) the maximum lag $K \in \{1, 2, 3\}$ bounds direct causal influence to at most 3 steps back, (2) geometric edge decay ($p_k = p \cdot 0.7^k$) makes distant lag connections increasingly sparse, and (3) bounded activations (e.g., $\tanh$, $\sin$) naturally dampen signal propagation. There is no explicit spectral radius constraint on the weight matrices; instead, stability is enforced empirically via value clipping to $[-1000, 1000]$ at each simulation step and divergence rejection (any trajectory with $|x| > 10^6$ or NaN/Inf is discarded and resampled). A 50-step burn-in period is discarded to approximate the stationary distribution.

This design implies that causal effects attenuate rapidly---typically within 1--5 steps after intervention onset. While this is appropriate for systems with short-range temporal dependencies, it may underweight long-range causal effects (e.g., economic policy impacts over months). A PFN trained on this prior would inherit this short-range bias.

\paragraph{Markovian noise and identifiability.}
The noise $\epsilon_t^{(i)}$ is sampled i.i.d.\ at each time step---independent across both time and variables. This Markovian assumption implies that the causal graph $\Gcal$ fully mediates all temporal dependencies: there are no hidden common causes acting across time beyond what the lagged edges encode. Identifiability within the modeled lag window follows from the additive noise model (ANM) structure \citep{peters2014causal}: each mechanism has the form $\sigma(\sum w_j x_j + b) + \epsilon$, and the use of non-Gaussian noise families (Uniform, Laplace) for non-root nodes supports distinguishing causal from non-causal associations under standard ANM identifiability results.

An extension to \emph{non-Markovian confounding}---where $\text{Cov}(\epsilon_t^{(i)}, \epsilon_{t-s}^{(j)}) \neq 0$ for some $s > 0$---would model latent confounders that persist over time (e.g., unobserved trends or regime states not captured by the graph). This would break the assumption that interventional distributions depend only on the explicit causal structure and is an important direction for increasing the prior's realism.

\paragraph{Interventional vs.\ counterfactual semantics.}
CausalTimePrior generates \emph{interventional} paired data, not \emph{counterfactual} paired data. Concretely, when \texttt{generate\_pair} produces an observational series $\mathbf{X}^{\text{obs}}$ and an interventional series $\mathbf{X}^{\text{int}}$, the two simulations draw \textbf{independent noise realizations} $\{\epsilon_t^{(i)}\}$ and $\{\tilde{\epsilon}_t^{(i)}\}$. This means the trajectories may differ even \emph{outside} the intervention window---not because of any causal effect, but because they are driven by different exogenous noise.

This corresponds to the \textbf{interventional} query $P(\mathbf{X} \mid \doop(X^{(i)}_{t_I} = c))$: ``what would a \emph{new} draw from the system look like under this intervention?'' The alternative is the \textbf{counterfactual} query $P(\mathbf{X}^{CF} \mid \mathbf{X}^{\text{obs}}, \doop(X^{(i)}_{t_I} = c))$: ``given \emph{this specific} observational realization, what \emph{would have happened} if we had intervened?'' Counterfactual inference requires the three-step abduction-action-prediction procedure \citep{pearl2009causality}: (1) infer the noise $\epsilon = \mathbf{X}^{\text{obs}} - f(\Pa)$ (abduction), (2) modify the structural equation for the intervened variable (action), and (3) re-simulate with the \emph{same} noise (prediction).

In our framework, generating counterfactual pairs would require sharing the pre-sampled noise tensor between the observational and interventional simulation runs, so that the two trajectories are identical before the intervention onset and diverge only through causal propagation of the intervention effect. This is a natural extension that would enable training PFNs for counterfactual estimation---a strictly harder task than interventional prediction, since it requires reasoning about unit-level rather than population-level effects.

\paragraph{Noise model details.}
All noise is additive (added after the nonlinear activation) and exogenous (independent of parent values). Root nodes (no parents in $G_0$) receive Gaussian noise with $\text{std} \sim \text{ShiftedExp}(\text{rate}=1.0, \text{shift}=0.1)$, providing larger driving noise. Non-root nodes receive noise from one of three variance-matched families (each with probability $\tfrac{1}{3}$): Gaussian $\mathcal{N}(0, \text{std}^2)$, Uniform $U(-a, a)$ with $a = \text{std}\sqrt{3}$, or Laplace $\text{Lap}(0, b)$ with $b = \text{std}/\sqrt{2}$, where $\text{std} \sim \text{ShiftedExp}(\text{rate}=10.0, \text{shift}=0.01)$. The smaller non-root noise ensures that most of a variable's variance comes from its causal parents rather than its own noise term.

\paragraph{Stationarity scope.}
Within a single SCM instance, mechanisms are stationary: weights, biases, activation functions, and noise distributions are fixed across all time steps. Non-stationarity enters only through regime-switching SCMs (15\% of samples), where 2--3 regimes---each with its own causal graph and mechanisms---alternate according to a sticky Markov chain. The current framework does not support continuously time-varying coefficients, one-time structural breaks, or heteroscedastic noise. These extensions would broaden the prior's coverage of real-world non-stationary dynamics.

\paragraph{PFN prediction horizon.}
During training, 70\% of queries target a different variable 1--5 steps after intervention onset (downstream causal effects), while 30\% query the intervention target itself at the intervention time (instantaneous effects). The PFN architecture accepts arbitrary $(\text{intervention\_time}, \text{query\_time})$ pairs as normalized inputs and can in principle predict at any horizon, but accuracy is expected to degrade beyond the 0--5 step training concentration. Extending the training distribution to longer horizons---potentially with curriculum learning \citep{bengio20009curriculum}---could improve long-range causal effect estimation.

\section{Intervention Type Examples}
\label{app:intervention_examples}

\begin{figure}[ht]
\centering
\includegraphics[width=0.95\textwidth]{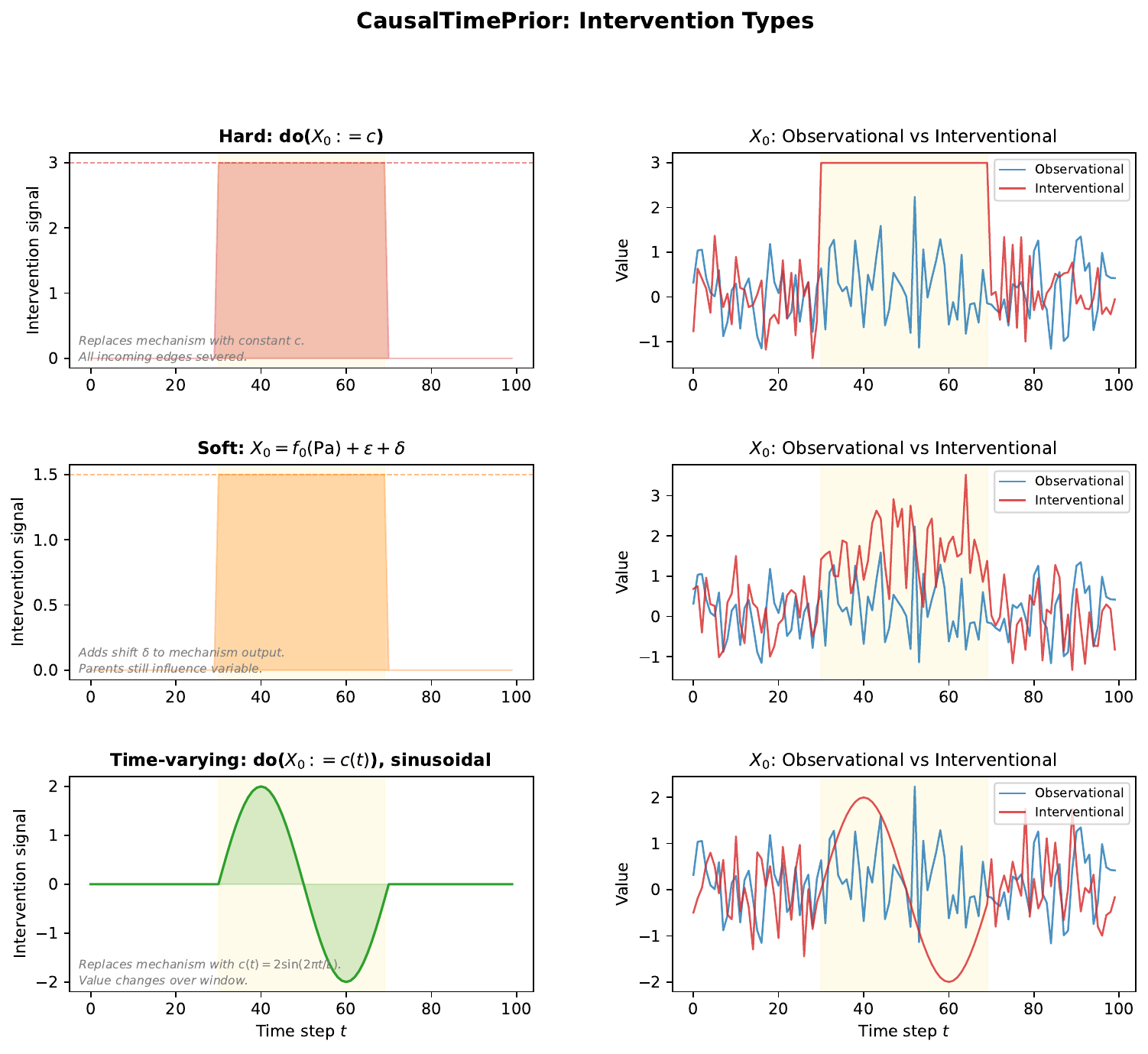}
\caption{The three intervention types in CausalTimePrior. Left column: the intervention signal applied to $X_0$. Right column: resulting observational (blue) vs.\ interventional (red) trajectories. Yellow shading marks the intervention window $[30, 70)$.}
\label{fig:intervention_types}
\end{figure}

Figure~\ref{fig:intervention_types} illustrates the three intervention types supported by CausalTimePrior, applied to the same 10-variable TSCM. Each row shows one type: (1)~\textbf{Hard intervention} ($\doop(X_0 := 3.0)$) replaces the variable's mechanism entirely with a constant, severing all incoming causal edges during the intervention window---visible as a flat interventional trajectory. (2)~\textbf{Soft intervention} ($\delta = 1.5$) adds an additive shift to the mechanism output while preserving parental influence---the interventional trajectory tracks the observational one but is displaced. (3)~\textbf{Time-varying intervention} ($\doop(X_0 := 2\sin(2\pi t/L))$) replaces the mechanism with a time-dependent function, producing a sinusoidal interventional trajectory.

\begin{figure}[ht]
\centering
\includegraphics[width=0.95\textwidth]{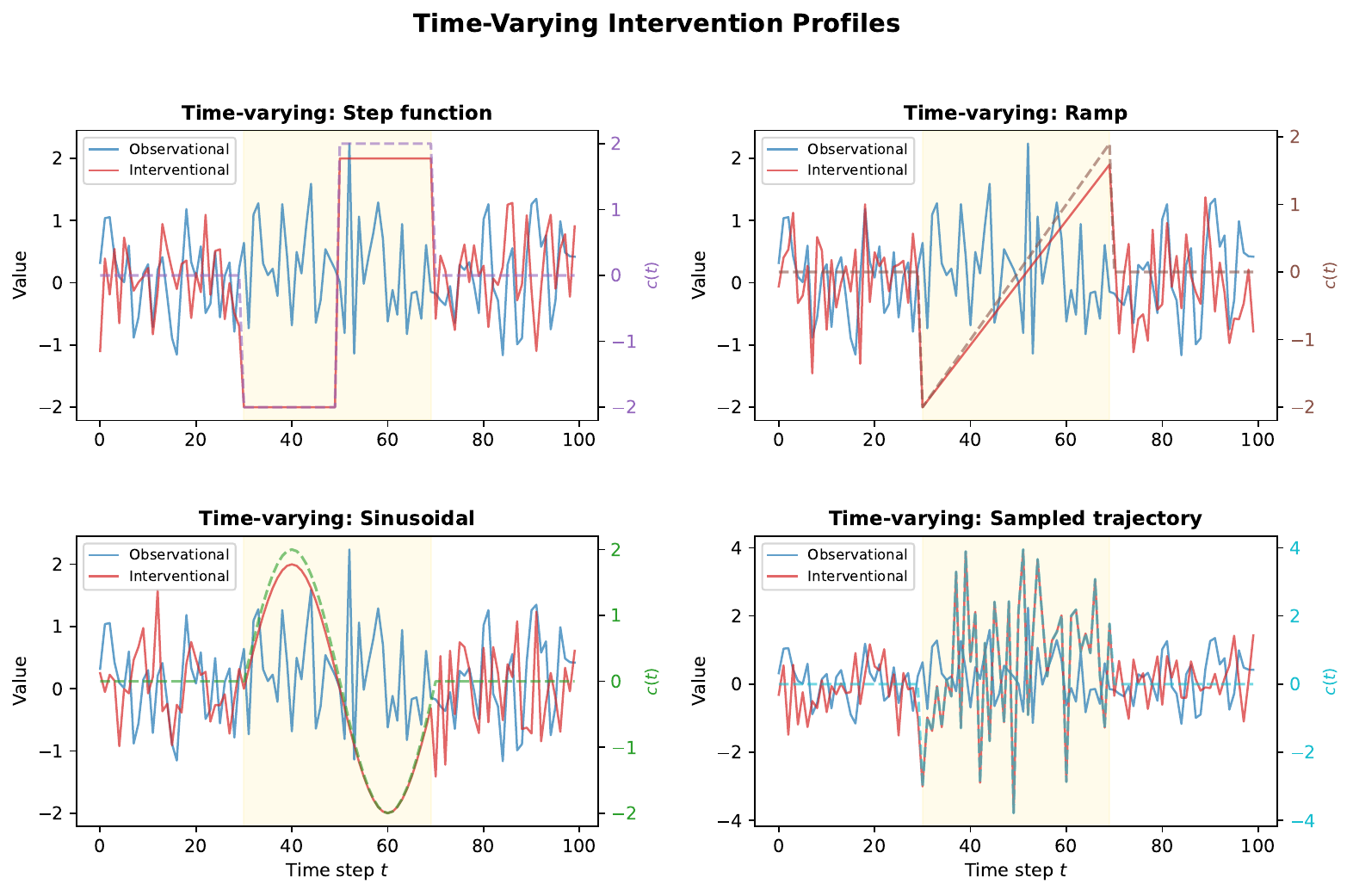}
\caption{Four time-varying intervention profiles. Blue: observational trajectory. Red: interventional trajectory. Dashed: intervention signal $c(t)$ (right axis). All four profiles are applied to the same variable in the same TSCM.}
\label{fig:intervention_profiles}
\end{figure}

Figure~\ref{fig:intervention_profiles} shows the four time-varying intervention profiles in detail, each applied to the same SCM and variable. The \textit{step} profile jumps from $-2$ to $+2$ at the midpoint; the \textit{ramp} linearly interpolates from $-2$ to $+2$; the \textit{sinusoidal} profile follows $c(t) = 2\sin(2\pi t/L)$; and the \textit{sampled trajectory} draws independent $c(t) \sim \mathcal{N}(0,4)$ at each time step. The dashed line shows the intervention signal $c(t)$ on the secondary axis.

\section{Example Visualizations}
\label{app:visualizations}

\begin{figure}[ht]
\centering
\includegraphics[width=0.9\textwidth]{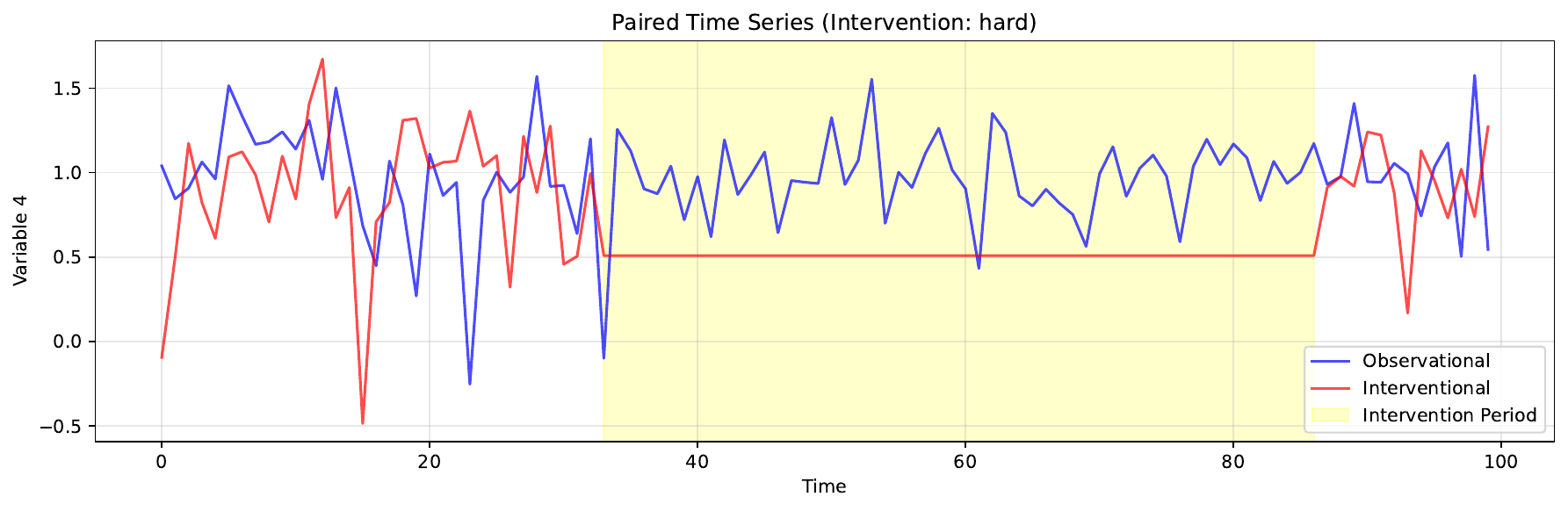}
\caption{Paired observational and interventional time series for the intervention target variable. The yellow shaded region indicates the intervention period. The divergence between blue (observational) and red (interventional) trajectories demonstrates the causal effect of the intervention.}
\label{fig:paired_timeseries}
\end{figure}

\begin{figure}[ht]
\centering
\includegraphics[width=0.95\textwidth]{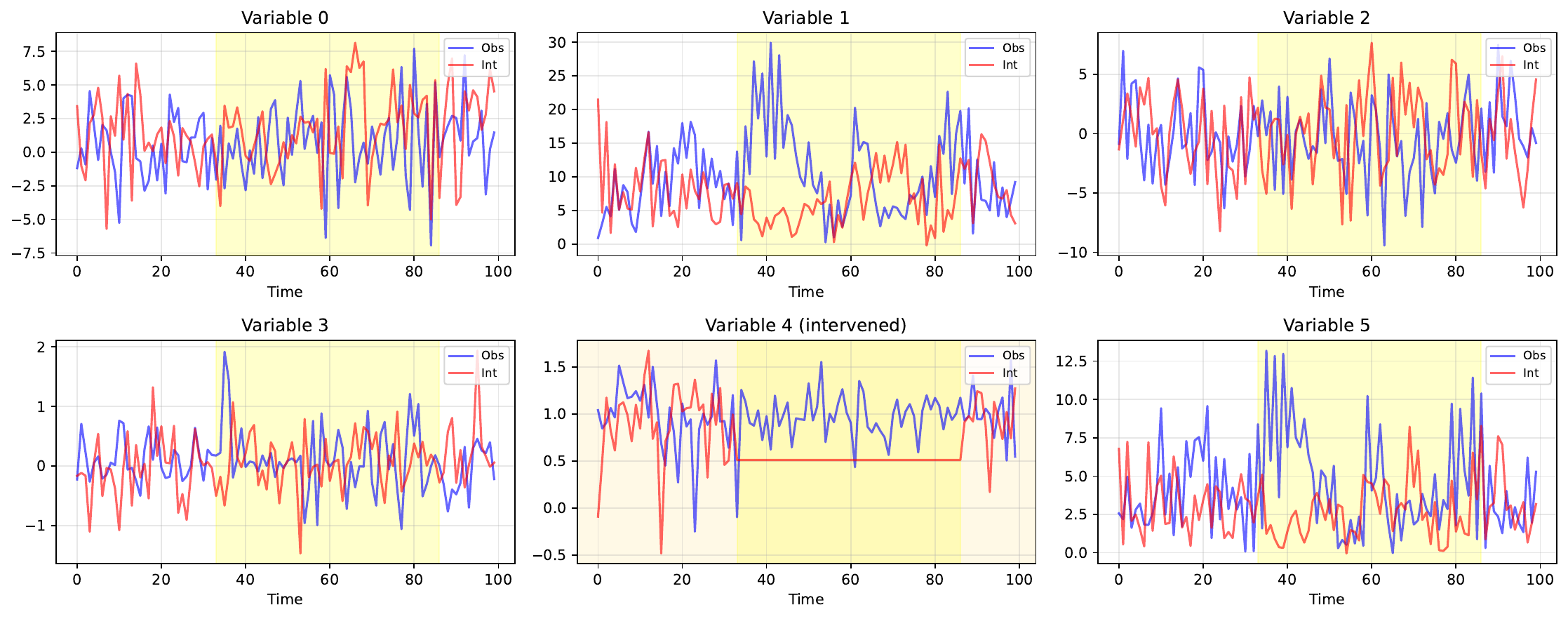}
\caption{All variables in a sampled 6-variable TSCM with a hard intervention on Variable 4. The intervention target is highlighted with a yellow background. Causal effects propagate through the graph structure, affecting downstream variables while leaving non-causally connected variables unchanged.}
\label{fig:all_variables}
\end{figure}

Figure~\ref{fig:paired_timeseries} shows an example of paired observational and interventional time series generated from CausalTimePrior. The hard intervention on Variable 4 between $t=20$ and $t=80$ causes a clear divergence between the observational (blue) and interventional (red) trajectories during and after the intervention period. Figure~\ref{fig:all_variables} displays all 6 variables in the sampled TSCM, with Variable 4 (the intervention target) highlighted. The propagation of intervention effects through the causal graph is visible in downstream variables, while non-causally connected variables remain unaffected.

\section{Prior Property Distributions}
\label{app:prior_distributions}

\begin{figure}[ht]
\centering
\includegraphics[width=0.95\textwidth]{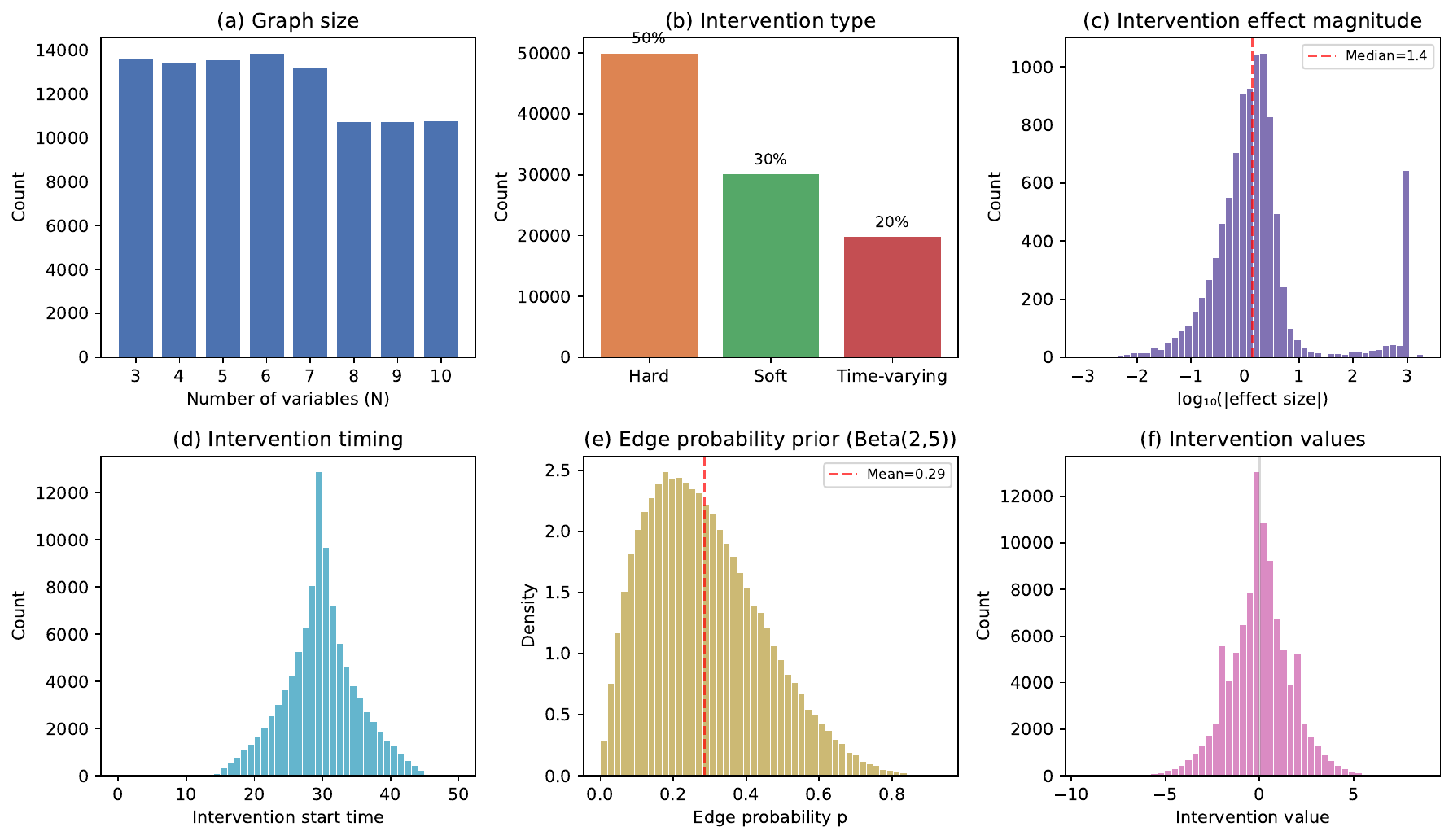}
\caption{Distributions of prior properties across 100K sampled TSCMs from CausalTimePrior. The prior produces diverse graph structures, intervention types, and effect magnitudes.}
\label{fig:prior_distributions}
\end{figure}

Figure~\ref{fig:prior_distributions} shows the distributions of key properties across 100K TSCMs sampled from CausalTimePrior. (a)~Graph sizes are approximately uniform over $N \in [3, 10]$. (b)~Intervention types are split 50\% hard, 30\% soft, and 20\% time-varying. (c)~Intervention effect magnitudes span several orders of magnitude (median 1.4) on a log scale, ensuring the prior covers both subtle and large causal effects. (d)~Intervention start times are concentrated in the second half of the sequence to allow sufficient observational context. (e)~Edge probabilities follow a Beta(2,5) prior (mean 0.29), producing mostly sparse graphs. (f)~Intervention values are approximately Gaussian-distributed around zero.

\section{Implementation Details}
\label{app:details}

We implement a simple proof-of-concept architecture using a 2-layer GRU encoder (128 hidden dim). The \textbf{temporal encoder} processes the observational time series $\mathbf{X}^{\text{obs}}_{1:T}$, the \textbf{intervention encoder} embeds the intervention specification $\doop(X^{(i)}_t = c)$ (which variable, when, what value), and the \textbf{query encoder} embeds the prediction query (which variable to predict, when). The \textbf{prediction head} combines these encodings to output $P(Y_\tau^{\text{int}} | \doop(X^{(i)}_t = c), \mathbf{X}^{\text{obs}}_{1:T})$ as a Gaussian distribution with predicted mean and standard deviation. 
The training objective is:
\begin{equation}
    \mathcal{L}(\theta) = \E_{\Scal \sim \Pi} \E_{\mathbf{X}^{\text{obs}}_{1:T} \sim P_{\text{obs}}^\Scal} \E_{(x_t, y_\tau) \sim P_{\text{int}}^\Scal} \left[ -\log q_\theta(y_\tau | \doop(x_t), \mathbf{X}^{\text{obs}}_{1:T}) \right]
\end{equation}
where $q_\theta$ is a Gaussian with predicted mean and variance, $\mathbf{X}^{\text{obs}}_{1:T}$ is the observational time series context, $\doop(x_t)$ is the temporal intervention query at time $t$, and $y_\tau$ is the interventional outcome at target time $\tau$.

\paragraph{Prior hyperparameters.} $N_{\max} = 10$, $K_{\max} = 3$, $\alpha = 2$, $\beta = 5$ (sparse graphs), $\gamma = 0.7$ (lag decay), $\sigma_w = 1.0$, $\sigma_b = 0.5$.

\paragraph{Training.} Adam optimizer, learning rate $10^{-4}$, batch size 32, sequence length 50, 15 epochs on 100K TSCMs. Training takes approximately 11 minutes on CPU (Intel/AMD with AVX2). The model checkpoint requires ~330KB of storage.

\paragraph{Architecture.} We use a simplified 2-layer GRU encoder with 128 hidden dimensions and Gaussian prediction head (mean + standard deviation). We chose a GRU over transformer architectures (as used in Do-PFN and TimePFN) for computational simplicity in this proof-of-concept: the GRU processes variable-length sequences efficiently and its recurrent state naturally captures temporal dependencies. This choice prioritizes fast iteration ($\sim$11 min training on CPU) over architectural optimality; scaling to transformer or GatedDeltaProduct architectures \citep{moroshan2025tempopfn} is an important next step.

\paragraph{Implementation.} CausalTimePrior is implemented from scratch, drawing conceptual inspiration from TempoPFN's diverse generator design and intervention logic from Do-PFN. The core is a \texttt{TemporalSCM} class that supports both \texttt{sample\_observational(T)} and \texttt{sample\_interventional(T, intervention)} methods, enabling paired data generation from the same underlying causal structure with time-lagged dependencies and multiple intervention types.

\section{Results}\label{app:results}

We train a 2-layer GRU encoder (128 hidden dim) for 15 epochs on 100K TSCMs and evaluate on 1,000 held-out TSCMs using three query types: (1) \textbf{Intervened} (query the intervention target itself), (2) \textbf{Downstream} (query a variable causally reachable from the intervention), and (3) \textbf{Non-causal} (query a variable with no causal path from the intervention).

Table~\ref{tab:results} shows the model's predictions are most accurate for intervened queries (Pred/GT = 0.95), reasonable for downstream queries (0.85), and substantially biased downward for non-causal queries (0.46). The low non-causal ratio reflects the model correctly predicting near-zero causal effect for non-causal variables, combined with nonzero ground-truth values at those time steps due to the variables' own dynamics. The model predicts $\sim$2$\times$ larger effects for causally connected queries compared to non-causal queries (33.91 vs 15.66 mean prediction).

\paragraph{Baselines.} We compare against a Vector Autoregression (VAR-OLS) fitted per-dataset, PCMCI+ \citep{runge2020pcmci} which discovers causal graphs via conditional independence tests, and a mean prediction baseline (Table~\ref{tab:baselines}). SimpleCausalPFN achieves comparable RMSE to VAR-OLS (176.4 vs 176.5) while requiring no per-dataset fitting. PCMCI+ achieves lower overall RMSE (161.4) by leveraging discovered causal structure, but requires expensive per-sample graph discovery.

\paragraph{Shuffled-intervention control.} To test whether the model's query-type sensitivity reflects learned causal structure rather than distributional artifacts, we evaluate with randomly shuffled intervention targets. Under shuffling, predictions for intervened queries change substantially (+33\% mean prediction) and Pred/GT degrades from 0.95 to 1.26, indicating the model is sensitive to intervention target information. However, non-causal predictions remain low (+13\%), suggesting the model has also partially learned distributional properties of variable positions, motivating richer architectures that more explicitly encode causal graph structure.

\begin{table}[ht]
\caption{Three-way evaluation on held-out TSCMs. NMSE (Normalized MSE = MSE/Var(GT); NMSE$<$1 is better than predicting the mean). The model's Pred/GT ratio is highest for intervened queries (0.95) and lowest for non-causal queries (0.46), suggesting learned causal understanding. Overall NMSE $\approx$ 1.0 indicates limited absolute prediction quality.}
\label{tab:results}
\centering
\small
\vskip 0.1in
\begin{tabular}{@{}lcccccc@{}}
\toprule
\textbf{Query Type} & \textbf{Queries} & \textbf{RMSE} $\downarrow$ & \textbf{NMSE} $\downarrow$ & \textbf{Mean Pred} & \textbf{Mean GT} & \textbf{Pred/GT} $\uparrow \downarrow = 1$ \\
\midrule
Intervened & 573 & 226.39 & 1.41 & 33.91 & 35.68 & \textbf{0.95} \\
Downstream & 270 & 231.31 & 0.67 & 57.50 & 67.70 & 0.85 \\
Non-causal & 157 & \textbf{143.15} & \textbf{0.66} & 15.66 & 33.99 & 0.46 \\
\midrule
Overall & 1000 & 216.87 & 0.99 & 37.41 & 44.06 & 0.85 \\
\bottomrule
\end{tabular}
\end{table}

\begin{table}[ht]
\caption{Comparison with baselines on held-out TSCMs. SimpleCausalPFN achieves comparable RMSE to VAR-OLS while requiring no per-dataset fitting. PCMCI+ achieves the lowest RMSE but requires per-sample causal graph discovery.}
\label{tab:baselines}
\centering
\small
\vskip 0.1in
\begin{tabular}{@{}lcccc@{}}
\toprule
\textbf{Method} & \textbf{RMSE} $\downarrow$ & \textbf{MAE} $\downarrow$ & \textbf{Effect Dir. Acc.} $\uparrow$ & \textbf{Effect Size Corr.} $\uparrow$ \\
\midrule
Oracle & 0.00 & 0.00 & 100.0\% & 1.000 \\
\midrule
PCMCI+ & \textbf{161.38} & \textbf{28.67} & 74.1\% & 0.784 \\
SimpleCausalPFN & 176.45 & 34.92 & 70.4\% & \textbf{0.821} \\
VAR-OLS & 176.45 & 33.86 & \textbf{93.7\%} & \textbf{0.821} \\
Mean Prediction & 256.18 & 73.26 & 60.4\% & 0.286 \\
\bottomrule
\end{tabular}
\end{table}

\section{Intervention Type Ablation}
\label{app:ablation}

We investigate whether training on diverse intervention types (hard, soft, time-varying) improves performance compared to training only on hard interventions. Table~\ref{tab:ablation} compares the mixed-intervention model (trained on 100K TSCMs with all intervention types) against a hard-only model (trained on 10K TSCMs with only hard interventions) on the same three-way test set.

\begin{table}[ht]
\caption{Intervention type ablation. The mixed-intervention model achieves higher effect direction accuracy and effect size correlation compared to the hard-only model.}
\label{tab:ablation}
\centering
\small
\vskip 0.1in
\begin{tabular}{@{}lcccc@{}}
\toprule
\textbf{Model} & \textbf{RMSE} $\downarrow$ & \textbf{MAE} $\downarrow$ & \textbf{Effect Dir. Acc.} $\uparrow$ & \textbf{Effect Size Corr.} $\uparrow$ \\
\midrule
Mixed (100K) & 216.87 & 81.68 & \textbf{70.4\%} & \textbf{0.821} \\
Hard-only (10K) & \textbf{212.16} & \textbf{69.68} & 63.9\% & 0.691 \\
\bottomrule
\end{tabular}
\end{table}

The mixed model achieves higher effect direction accuracy (70.4\% vs 63.9\%) and effect size correlation (0.821 vs 0.691), suggesting that diversity in intervention types during training improves the model's ability to reason about causal effects. While the hard-only model achieves slightly lower overall RMSE (212.16 vs 216.87), this is likely due to the simpler prediction task when only hard interventions are present.

\section{Out-of-Distribution Generalization}
\label{app:ood}

We evaluate whether the model generalizes to TSCMs with structural properties outside the training distribution. We generate an out-of-distribution (OOD) test set of 1,000 TSCMs with: (1) larger graphs ($N \in [8,10]$ vs training mean $\sim$6), (2) maximum lag $K=3$, (3) denser graphs (edge probability $\geq 0.3$ vs training mean $\sim$0.29), and (4) only complex nonlinear mechanisms (sin, cos, square, tanhReLU).

\begin{table}[ht]
\caption{Out-of-distribution generalization. Performance degrades on OOD TSCMs with larger, denser graphs and complex mechanisms, but the model retains basic causal structure (downstream RMSE $>$ intervened RMSE).}
\label{tab:ood}
\centering
\small
\vskip 0.1in
\begin{tabular}{@{}lccccc@{}}
\toprule
\textbf{Test Set} & \textbf{RMSE} $\downarrow$ & \textbf{NMSE} $\downarrow$ & \textbf{MAE} $\downarrow$ & \textbf{Effect Dir. Acc.} $\uparrow$ & \textbf{Effect Size Corr.} $\uparrow$ \\
\midrule
In-distribution & \textbf{216.87} & 0.99 & \textbf{81.68} & \textbf{70.4\%} & \textbf{0.821} \\
OOD & 265.97 & \textbf{0.72} & 163.25 & 62.7\% & 0.599 \\
\bottomrule
\end{tabular}
\end{table}

As expected, performance degrades on OOD data: RMSE increases from 216.87 to 265.97 and effect size correlation drops from 0.821 to 0.599. The lower OOD NMSE (0.72 vs 0.99) reflects higher ground-truth variance in the OOD test set rather than better prediction quality. However, the model retains some causal understanding, with intervened queries (RMSE=237.01) outperforming downstream queries (RMSE=313.92), consistent with the in-distribution pattern.

\section{Causal vs.\ Correlation-Based Prediction}
\label{app:causal_example}

To illustrate how the PFN leverages causal structure rather than correlations, consider a concrete test case. In sample 626 from our test set, a 3-variable temporal TSCM is intervened on variable 2 at $t=25$, and we query variable 0 at $t=28$. There is no causal path from variable 2 to variable 0, but the observational time series exhibit a correlation of $-0.49$ between them. The ground truth interventional value ($-0.056$) is nearly identical to the observational baseline ($-0.094$), confirming a near-zero causal effect (0.039). The PFN predicts $-0.050$ (error $0.005$), correctly recognizing the absence of causal influence despite the spurious correlation. In contrast, VAR-OLS---which relies on learned correlations without distinguishing causal from non-causal associations---predicts $-0.992$ (error $0.936$), a $177\times$ larger prediction error. This pattern generalizes: across 157 non-causal queries, the PFN achieves lower prediction error than VAR-OLS in 45\% of cases, with the largest gains precisely on samples exhibiting high spurious correlations ($|\rho| > 0.3$).

\end{document}